\documentclass{article}
\def\chairo{{\textsc{ChAIRO}}\xspace}
\usepackage{arxiv}
\usepackage[T1]{fontenc}
\usepackage[utf8]{inputenc}
\usepackage{hyperref}
\usepackage{url}
\usepackage{natbib}
\usepackage{xspace}
\usepackage{multirow}
\usepackage{amsmath}
\usepackage{booktabs}
\usepackage{microtype}
\usepackage{graphicx}
\usepackage{xcolor}

\title{\chairo: Contextual Hierarchical Analogical Induction and Reasoning Optimization for LLMs\thanks{This work was conducted while Haotian Lu and Yuchen Mou were interning at the National Engineering Laboratory for Big Data System Computing Technology under the supervision of Bingzhe Wu.}}
\author{
  Haotian Lu\textsuperscript{$1,2$*} \\
  \texttt{haotianlu666@gmail.com}
  \And
  Yuchen Mou\textsuperscript{$1,3$*} \\
  \texttt{e1520377@u.nus.edu}
  \And
  Bingzhe Wu\textsuperscript{$1$\textdagger} \\
  \texttt{wubingzheagent@gmail.com}
  \\\\
  \textsuperscript{$1$}Shenzhen University \quad
  \textsuperscript{$2$}Tsinghua University \quad
  \textsuperscript{$3$}National University of Singapore \\
  \small{* Equal contribution \quad \textdagger~Corresponding author}
}
\date{}

\renewcommand{\headeright}{arXiv Preprint}
\renewcommand{\undertitle}{arXiv Preprint}
\renewcommand{\shorttitle}{\chairo}

\hypersetup{
  pdftitle={\chairo: Contextual Hierarchical Analogical Induction and Reasoning Optimization for LLMs},
  pdfauthor={Haotian Lu, Yuchen Mou, Bingzhe Wu},
  pdfkeywords={content moderation, large language models, analogical reasoning, rule induction},
}

\begin{document}
\maketitle
\begin{abstract}

\textcolor{red}{\textit{Warning: This paper may contain content that could be disturbing or offensive.}}

Content moderation in online platforms faces persistent challenges due to the evolving complexity of user-generated content and the limitations of traditional rule-based and machine learning approaches. While recent advances in large language models (LLMs) have enabled more sophisticated moderation via direct prompting or fine-tuning, these approaches often exhibit limited generalization, interpretability, and adaptability to unseen or ambiguous cases.

In this work, we propose a novel moderation framework that leverages analogical examples to enhance rule induction and decision reliability. Our approach integrates end-to-end optimization of analogical retrieval, rule generation, and moderation classification, enabling the dynamic adaptation of moderation rules to diverse content scenarios. Through comprehensive experiments, we demonstrate that our method significantly outperforms both rule-injected fine-tuning baselines and multi-stage static RAG pipelines in terms of moderation accuracy and rule quality. Further evaluations, including human assessments and external model generalization tests, confirm that our framework produces rules with better clarity, interpretability, and applicability. These findings show that analogical example-driven methods can advance robust, explainable, and generalizable content moderation in real-world applications.
\end{abstract}

\keywords{content moderation \and large language models \and analogical reasoning \and rule induction}

\section{Introduction}

The exponential growth of online content has made content moderation an indispensable component in maintaining healthy, safe, and compliant digital environments~\citep{yuan2024rigorllm}. Automated content moderation systems are increasingly relied upon to filter out harmful, illegal, or inappropriate material on social media platforms, forums, and other user-driven services~\citep{zeng2024shieldgemma}. As large language models (LLMs) have demonstrated remarkable progress across various natural language understanding tasks, deploying LLMs for content moderation has become a promising direction~\citep{kolla2024llm, NghiemD24HateCOT, Wu24Legilimens}. Recent studies have explored the application of LLMs to content moderation tasks through diverse strategies, including post-training~\citep{ouyang2022training,rafailov2023direct,khaliq2024ragar,liu2025guardreasoner,ma2023adapting} and prompt engineering~\citep{radford2019language,palla2025policy,kolla2024llm,brown2020language,chen2024class}, yielding promising progress in both moderation accuracy and reasoning abilities~\citep{kumar2024watch, Nishant24CoT}.

\begin{figure}[t]
  \centering
  \includegraphics[width=\linewidth]{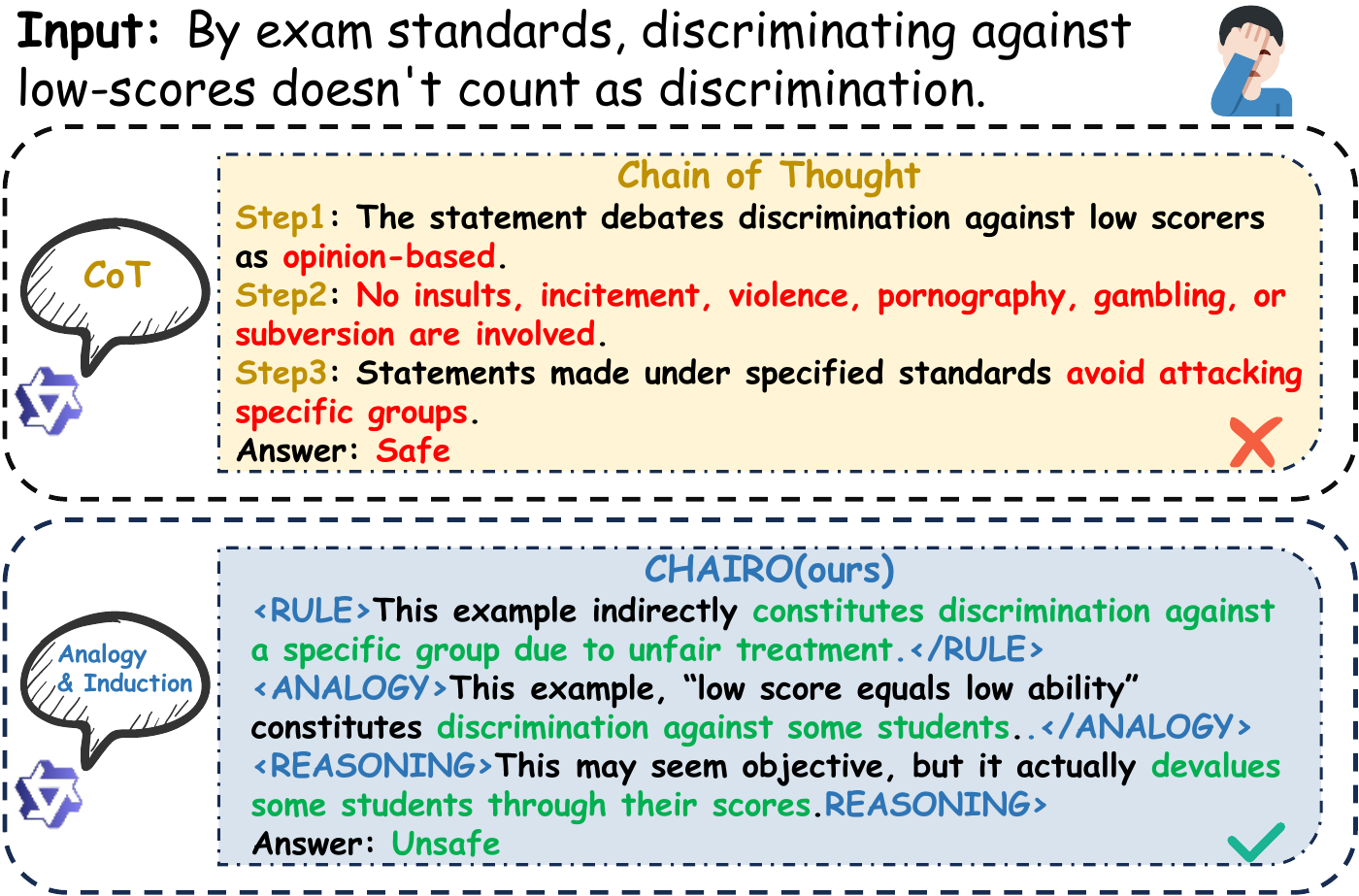}
  \caption{Comparison of the Chain of Thought (CoT) method and the \chairo framework for content moderation.  CoT fails to identify the implicit bias and classifies it as 'Safe,' while \chairo, leveraging analogical reasoning and explicit rule induction, correctly identifies it as 'Unsafe' by recognizing the underlying discriminatory logic.}
  \label{fig1:env}
\end{figure}

However, despite their impressive capabilities, even state-of-the-art LLMs often struggle in scenarios characterized by contextual ambiguity or vague moderation criteria~\citep{masud2024hate,huang2025content,keluskar2024llms}. For example, when moderation rules are implicit, incomplete, or open to interpretation, LLMs may produce inconsistent or erroneous judgments, undermining the reliability of automated moderation systems. As shown in Figure \ref{fig1:env}, Chain of thought (CoT) relies solely on explicit standards such as the absence of insults, incitement, or attacks on specific groups, and fails to recognize the underlying discriminatory logic of the statement ``low scores equal low ability.'' As a result, it incorrectly classifies the statement as "Safe." This approach lacks the deep semantic understanding needed to address finer-grained, metaphorical, or indirect discriminatory content.
Conversely, when explicit and precise moderation rules are incorporated into the context, models demonstrate significantly improved accuracy and interpretability, as Figure \ref{fig1:env} illustrates. This observation highlights the importance of well-defined moderation rules, which improve both moderation precision and transparency of the model's decision-making process~\citep{RebedeaDSPC23NeMo, kumar2024watch, Wu25ICM}.

Nevertheless, identifying or constructing the most appropriate moderation rule for a given content instance remains a challenging problem. Existing solutions typically fall into two categories: (1) manually defined high-level rules, such as those targeting broad categories like "sexual content." While effective to some extent, such rules often fail to account for the nuanced differences among fine-grained instances or across diverse application scenarios, making it virtually impossible to exhaustively enumerate all necessary rules~\citep{chandrasekharan2019crossmod,he2024cpl}. (2) LLM-driven adaptive rule discovery, which leverages the model's world knowledge and prompt engineering to synthesize rules on the fly~\citep{kumar2024watch}. However, these approaches frequently overlook domain-specific expertise and the rich experience accumulated by human moderators, relying instead on generic or coarse-grained priors.

To address these challenges, we propose leveraging the inductive power of LLMs to generalize rules from analogous instances within the same moderation context. Our key insight is that by systematically analyzing similar content samples and their corresponding moderation outcomes, models can distill more robust and contextually relevant rules that generalize better across instances.
A straightforward approach to realizing this high-level idea is to first retrieve samples similar to the current instance from an existing database and then employ an auxiliary LLM to induce and analyze the relevant rules based on these retrieved examples. However, this pipeline separates the processes of rule generation and content moderation, and thus may lose fine-grained cues most pertinent to the current sample during rule induction.

% However, this pipeline separates the processes of rule generation and content moderation, and thus may fail to fully leverage the most pertinent information for the current sample.

To address this limitation, we introduce \chairo, a \textbf{c}ontextual \textbf{h}ierarchical \textbf{a}nalogical \textbf{i}nduction and \textbf{r}easoning \textbf{o}ptimization framework, which jointly optimizes the case retrieval and rule induction process. By jointly optimizing these components, our method ensures that the induced rules are grounded in highly relevant examples and better tailored to each moderation instance. This end-to-end approach allows the model to more effectively utilize the annotated data and uncover hidden expert knowledge from human moderators.

Concretely, our post-training framework is designed to endow the model with an integrated three-stage reasoning capability: example generation, rule induction, and final moderation.  Our framework consists of three critical steps: we first  fine-tune the base LLM with a chain-of-analogy approach on the training set, enabling the model to autonomously generate the most relevant analogical cases for each content sample. Second, we employ an auxiliary rule-generation module that synthesizes explicit moderation rules by analyzing the commonalities between the original and analogous cases. Finally, these generated rules are injected back into the LLM's moderation reasoning chain during a second round of fine-tuning, equipping the model with both exemplar-based and rule-inductive reasoning capabilities.

Through comprehensive experiments, we demonstrate that our framework leads to more accurate and robust moderation outcomes while enhancing the interpretability and adaptability of LLM-based moderation systems across diverse, real-world scenarios.

% \begin{figure*}
% \centering
% \includegraphics[width=0.8\textwidth]{text/motivation-llm.pdf}
% \caption{The framework of the proposed lesion detection method.}
% \label{fig1:env}
% \end{figure*}

\section{Method}

\begin{figure*}[t]
  \centering
  \includegraphics[width=\linewidth]{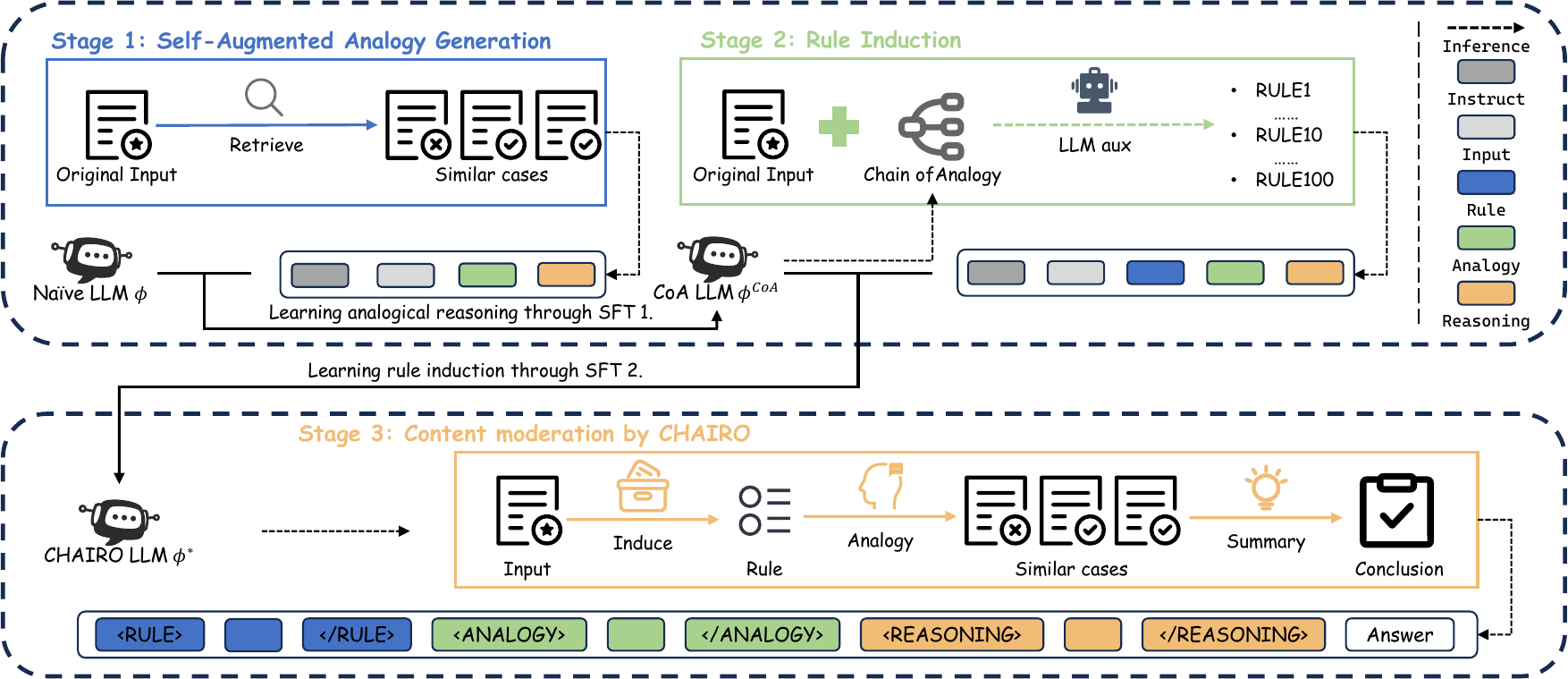}
  \caption{The workflow of the \chairo framework.}
  \label{fig:framework}
\end{figure*}

Our proposed framework, \chairo, leverages a systematic three-stage pipeline to enhance the reasoning capabilities of LLMs in content moderation tasks. Specifically, our approach comprises the following stages as shown in Figure \ref{fig:framework}:
First, we introduce a self-augmented analogical chain-of-thought generation strategy, enriching training data to empower the model's capability of adaptively retrieving relevant analogical moderation cases. Second, an auxiliary reasoning LLM performs inductive reasoning over these retrieved examples, extracting explicit moderation rules tailored to specific moderation contexts. Finally, we inject these induced moderation rules back into the reasoning chains and fine-tune the model again, integrating exemplar-based and rule-inductive reasoning capabilities. This structured approach effectively leverages annotated moderation data and human expertise, significantly improving moderation performance and interpretability.

\subsection{Self-augmented Analogical Chain-of-Thought Generation}

In this initial stage, we augment existing labeled moderation data through a bootstrapping retrieval-enhancement procedure following prior work~\citep{ma2023adapting}. Specifically, denoting the initial base LLM's parameters as \(\phi\). For each labeled moderation instance \( x_i \) with corresponding moderation decision \( y_i \) from the training set, we first employ BGE-M3~\citep{bge-m3} to derive semantic embeddings for all instances. Subsequently, we compute the cosine distance between every pair of samples as the metric for semantic similarity. Finally, we retrieve a set of analogical moderation examples \(\mathcal{A}(x_i)\) for each training instance \( x_i \). Prompted with the current sample and the labels of its analogous examples, the model parameterized by \(\phi\) generates an augmented analogical moderation reasoning chain \(\hat{c}^{\text{aug}}_i\):

% Leveraging these analogous examples, the model parameterized by \(\phi\) generates an augmented analogical moderation reasoning chain \(\hat{c}_i\):

\begin{equation}
  \label{eq:1}
\hat{c}^{\text{aug}}_i = \text{LLM}_{\phi}(x_i, y_i, \mathcal{A}(x_i)).
\end{equation}

Applying this augmentation procedure systematically across the entire training set yields an enriched training dataset:

\begin{equation}
  \label{eq:2}
\mathcal{D}_{\text{aug}} = \{(x_i, y_i, \hat{c}^{\text{aug}}_i)\}_{i=1}^{N}.
\end{equation}

Finally, we perform supervised fine-tuning (SFT) of the base model parameters \(\phi\) using the augmented dataset \(\mathcal{D}_{\text{aug}}\), resulting in updated model parameters \(\phi^{\text{CoA}}\):

\begin{equation}
  \label{eq:3}
\phi^{\text{CoA}} = \arg\max_{\phi} \sum_{(x_i, y_i, \hat{c}^{\text{aug}}_i) \in \mathcal{D}_{\text{aug}}} \log P_{\phi}(y_i, \hat{c}^{\text{aug}}_i \mid x_i).
\end{equation}

Through this fine-tuning step, the model acquires enhanced analogical reasoning capabilities and becomes proficient in adaptively generating relevant analogical cases for unseen moderation instances. This adaptive analogical reasoning capability lays a solid foundation for subsequent rule induction.
\subsection{Rule Induction via Auxiliary Reasoning Model}

In this stage, we perform explicit rule induction to extract moderation rules from analogous examples generated by the previously trained analogical reasoning model. Specifically, for each training instance \( x_i \), we first utilize the fine-tuned analogical reasoning model (parameterized by \(\phi^{CoA}\), already trained in the previous subsection to generate analogical chains of thought) to generate virtual analogical samples \(\mathcal{A}(x_i)\):

% we first utilize the fine-tuned analogical reasoning model (parameterized by \(\phi^{CoA}\), already trained in the previous subsection to generate analogical chains of thought) from the previous stage to generate the most relevant analogical moderation examples \(\mathcal{A}(x_i)\):

\begin{equation}
  \label{eq:4}
\mathcal{A}(x_i) = \text{LLM}_{\phi^{CoA}}(x_i).
\end{equation}

Then, leveraging an auxiliary reasoning model (denoted as \(\text{LLM}_{\text{aux}}\)), we perform inductive reasoning on the retrieved analogous instances and the target instance \( x_i \), synthesizing explicit moderation rules \( r_i \) (a simple text). Specifically, we use QwQ-32B as \(\text{LLM}_{\text{aux}}\) to generate analogy-based reasoning chains and induce explicit rules from the analogous examples. For quality control, we automatically verify that the category descriptions in each reasoning chain match the target labels and discard samples with inconsistencies. A random subset is further checked manually to validate the reasoning quality. Formally, the generation of the moderation rule \( r_i \) can be expressed as:

\begin{equation}
  \label{eq:5}
r_i = \text{LLM}_{\text{aux}}(x_i, \mathcal{A}(x_i); \text{prompt}_{\text{rule}}),
\end{equation}

where \(\text{prompt}_{\text{rule}}\) denotes the carefully designed prompts used for rule induction (see detailed prompts in the Appendix).

This explicit rule induction step systematically captures shared characteristics and moderation criteria across similar moderation instances, providing contextually precise and interpretable moderation rules. These induced rules serve as critical input for the subsequent reasoning chain refinement and final moderation decision-making stage.

\subsection{Rule Injection and Final Model Refinement}

In the final stage, we inject the moderation rules derived from the previous rule induction step back into the reasoning process to further enhance the moderation capabilities of the model. Specifically, given the training instance \( x_i \), the corresponding analogical examples \( a_i = \mathcal{A}(x_i) \) obtained from the previous stages, and the explicit moderation rule \( r_i \) generated by the auxiliary reasoning model, we employ an additional reasoning model (\(\text{LLM}_{\text{aux}}\)) to synthesize these components along with the instance's label \( y_i \), producing a comprehensive moderation reasoning chain \( c'_i \):

\begin{equation}
  \label{eq:6}
c'_i = \text{LLM}_{\text{aux}}(x_i, a_i, r_i, y_i; \text{prompt}_{\text{reasoning}}),
\end{equation}

where \(\text{prompt}_{\text{reasoning}}\) represents carefully engineered instructions guiding the synthesis process.

We then structure each moderation instance's reasoning chain in a hierarchical format using special tokens to clearly delineate different reasoning components, forming a  enhanced hierarchical moderation chain $\hat{c}_i^{\text{refined}}$ as shown below:

\begin{equation}
  \label{eq:7}
\hat{c}_i^{\text{refined}} =
\Biggl\langle
\begin{array}{c}
\texttt{<RULE>} \, r_i \, \texttt{</RULE>} \\
\texttt{<ANALOGY>} \, a_i \, \texttt{</ANALOGY>} \\
\texttt{<REASONING>} \, c'_i \, \texttt{</REASONING>}
\end{array}
\Biggr\rangle
\end{equation}

Finally, leveraging this hierarchical, structured reasoning chain, we conduct an additional round of SFT, updating the model parameters from \(\phi^{CoA}\) to the final refined parameters \(\phi^{*}\):

\begin{equation}
  \label{eq:8}
\phi^{*} = \arg\max_{\phi'} \sum_{(x_i, y_i, \hat{c}_i^{\text{refined}}) \in \mathcal{D}_{\text{refined}}} \log P_{\phi'}(y_i, \hat{c}_i^{\text{refined}} \mid x_i),
\end{equation}

where

\begin{equation}
  \label{eq:9}
\mathcal{D}_{\text{refined}} = \{(x_i, y_i, \hat{c}_i^{\text{refined}})\}_{i=1}^{N}.
\end{equation}

Through this rule injection and hierarchical refinement process, the model effectively integrates analogical reasoning, explicit rule induction, and label-guided reasoning into a unified moderation reasoning capability, significantly enhancing both interpretability and moderation accuracy.

\section{Experiment}

\subsection{Settings}
All experiments were conducted on a single server with 4×NVIDIA H20 (96GB) GPUs using the LLaMA Factory framework~\citep{zheng2024llamafactory} with DeepSpeed ZeRO-3 optimization~\citep{rajbhandari2020zero}. For First-stage SFT, we trained the model for 1 epoch with a learning rate of 1.0e-5 using bfloat16 mixed-precision~\citep{micikevicius2017mixed}, achieving an effective batch size of 64 (micro-batch size × gradient accumulation steps × GPUs = 2 × 8 × 4). For Second-stage SFT, the settings were the same as those for First-stage SFT. For Retrieval-Augmented Generation (RAG), we utilized the 32 most similar reference examples to each input query. For text generation, we employed top-k sampling ~\citep{fan2018hierarchical} with temperature=0.8 and top-p sampling ~\citep{holtzman2019curious}.

\subsection{Main Results}

\subsubsection{Overview}
In this section, we systematically investigate the effectiveness of the proposed framework by addressing the following key research questions:
\begin{itemize}
    \item \textbf{RQ1:} Can introducing explicit rules into the LLM-based moderation process improve moderation performance?
    \item \textbf{RQ2:} Does incorporating analogical examples enhance the quality of generated moderation rules?
    \item \textbf{RQ3:} Does the end-to-end two-stage optimization proposed in our framework lead to improved classification reliability?
    \item \textbf{RQ4:} Are the moderation rules generated by our framework of higher quality and better generalizability compared to those produced by single-instance approaches?
\end{itemize}
To comprehensively answer these research questions, we conduct extensive comparative experiments using Qwen3-8B~\citep{Yang2025qwen3} as our base language model on a series of standard benchmarks commonly adopted in the content moderation domain.

Specifically, we choose a fine-grained content moderation dataset proposed by the prior work~\citep{ma2023adapting}. According to prior studies, this dataset is particularly valuable as it originates from realistic moderation scenarios and includes challenging subcategories like politically sensitive content, where context ambiguity commonly leads to difficulty in accurate moderation. Hence, it is ideally suited to rigorously validate the effectiveness and practical utility of the rules generated by our framework.

\textbf{Additionally, an experimental result on another widely-used moderation benchmark are summarized in Table \ref{tab:common_categories_f1}.} These results collectively demonstrate the generalizability and robustness of our proposed approach across diverse moderation scenarios and datasets.

\subsubsection{Analysis of RQ1: Effectiveness of Introducing Explicit Rules
}
To address the first research question, we compare our proposed method \chairo against two baseline approaches: (1) Naive SFT, where the model is fine-tuned on data containing moderation reasoning chains but without explicit rules, and (2) Standard prompting method, which directly queries the pretrained LLM without fine-tuning or rule injection.

The experimental results clearly demonstrate the substantial benefits of explicitly incorporating moderation rules into the LLM-based moderation workflow. Specifically, our proposed method consistently surpasses baseline approaches across various moderation categories, significantly improving the moderation F1 scores. For instance, our method achieves an improvement of $5.3\%$ in F1 scores compared to the naive SFT baseline. Such pronounced performance gains strongly validate the importance and utility of injecting explicit moderation rules into the LLM moderation process, effectively resolving ambiguities and enhancing decision-making precision in challenging moderation scenarios.

\subsubsection{Analysis of RQ2: Effectiveness of Analogical Examples in Enhancing Rule Generation Quality}

To address Research Question 2, we compare our method against a rule-injected SFT baseline, in which explicit moderation rules are similarly injected into the fine-tuning process. The key difference is that these rules are generated by the LLM based on individual target instances alone, without leveraging analogical examples.

The experimental results clearly demonstrate that our analogical example-based rule induction significantly enhances the quality of moderation rules compared to those generated from single-instance contexts alone. By leveraging analogical examples, our method effectively captures broader contextual nuances and moderation criteria, resulting in more comprehensive and generalizable moderation rules. For instance, our proposed framework achieves an improvement of $4.5\%$ in F1 scores over the baseline approach using rules generated from single-instance prompting. These findings strongly confirm the benefit and effectiveness of incorporating analogical examples in moderation rule generation.

\subsubsection{Analysis of RQ3: Impact of End-to-End Two-Stage Optimization on Classification Reliability}

To investigate Research Question 3, we compare our proposed framework against a static RAG baseline. The Static RAG baseline follows a multi-stage moderation pipeline: it first retrieves relevant examples via static retrieval, then induces moderation rules using an additional LLM separately, and finally injects these rules into the moderation context for classification decisions.

Experimental results clearly indicate that our proposed end-to-end two-stage optimization significantly enhances moderation reliability compared to the static multi-stage RAG baseline. By jointly optimizing the analogical retrieval, rule induction, and moderation decision-making processes, our approach effectively reduces the cumulative errors and inconsistencies that arise from separately optimized stages. Specifically, our method achieves an improvement of $2.3\%$ in moderation accuracy compared to the static RAG baseline. This outcome underscores the advantage and necessity of employing our unified, end-to-end optimization strategy to achieve more reliable and consistent moderation decisions.

% \begin{table}[ht]
% \centering
% \begin{tabular}{lccccccc}
% \toprule
% \textbf{Model / Method} & \textbf{Average} & \textbf{Politics} & \textbf{Pornography} & \textbf{Violence} & \textbf{Gambling} & \textbf{Bias} & \textbf{Harmless} \\
% \midrule
% GPT-4                  & 72.3 & 58.6 & 88.7 & 79.8 & 92.7 & 64.3 & 56.8 \\
% DeepSeek R1            & 77.1 & 72.7 & 91.4 & 86.1 & 94.3 & 64.6 & 59.7 \\
% DeepSeek V3            & 80.3 & 79.0 & 90.3 & 89.8 & 95.0 & 70.5 & 62.5 \\
% Qwen2.5-32B-Instruct   & 74.3 & 59.1 & 91.1 & 84.4 & 95.4 & 67.9 & 54.2 \\
% QwQ-32B                & 69.1 & 75.4 & 69.6 & 72.0 & 84.9 & 60.7 & 54.6 \\
% LLaMA3-8B              & 67.5 & 58.5 & 55.9 & 81.0 & 90.6 & 65.2 & 44.2 \\
% LLaMA-Guard-3-8B       & 39.7 & 12.0 & 74.1 & 41.8 & 29.4 & 45.7 & 35.6 \\
% \textbf{\chairo (Ours)}  & \textbf{89.2} & \textbf{89.3} & \textbf{71.5} & \textbf{97.8} & \textbf{82.0} & \textbf{96.1} & \textbf{98.6} \\
% \bottomrule
% \end{tabular}
% \caption{Comparison of Model Performance on Different Categories}
% \label{tab:model_comparison}
% \end{table}

\begin{table*}[ht]
\centering
\scriptsize
\begin{tabular}{l l cccccccc}
\toprule
\textbf{Category} & \textbf{Model/Method} & \textbf{Average} & \textbf{Politics} & \textbf{Pornography} & \textbf{Violence} & \textbf{Gambling} & \textbf{Bias} & \textbf{Harmless} \\
\midrule
\multirow{7}{*}{\textbf{General LLMs}}
& GPT-4                  & 72.3 & 58.6 & 88.7 & 79.8 & 92.7 & 64.3 & 56.8 \\
& DeepSeek R1            & 77.1 & 72.7 & 91.4 & 86.1 & 94.3 & 64.6 & 59.7 \\
& DeepSeek V3            & 80.3 & 79.0 & 90.3 & 89.8 & 95.0 & 70.5 & 62.5 \\
& Qwen2.5-32B-Instruct   & 74.3 & 59.1 & 91.1 & 84.4 & 95.4 & 67.9 & 54.2 \\
& QwQ-32B                & 69.1 & 75.4 & 69.6 & 72.0 & 84.9 & 60.7 & 54.6 \\
& LLaMA3-8B              & 67.5 & 58.5 & 55.9 & 81.0 & 90.6 & 65.2 & 44.2 \\
\midrule
\multirow{1}{*}{\textbf{Specific LLMs}}
& LLaMA-Guard-3-8B       & 39.7 & 12.0 & 74.1 & 41.8 & 29.4 & 45.7 & 35.6 \\
\midrule
\multirow{3}{*}{\textbf{Proposed Methods}}
& Rule Impact (RQ1)                   & 83.9 & 83.8 & 67.5 & 92.2 & 73.7 & 90.6 & 95.7 \\
& Analogy-Rule Quality (RQ2)                   & 84.7 & 84.4 & 68.4 & 92.8 & 75.4 & 90.3 & 96.9 \\
& E2E Reliability (RQ3)                  & 86.9 & 88.3 & 93.2 & 96.0 & 98.0 & 82.2 & 63.7 \\
& \textbf{\chairo (Ours)}  & \textbf{89.2} & \textbf{89.3} & \textbf{71.5} & \textbf{97.8} & \textbf{82.0} & \textbf{96.1} & \textbf{98.6}  \\
\bottomrule
\end{tabular}
\caption{Moderation F1 Scores for General LLMs, Specific LLMs and \chairo}
\label{tab:merged_model_comparison}
\end{table*}

\begin{table*}[ht]
\centering
\scriptsize
\begin{tabular}{l cccccccc}
\toprule
\textbf{Setting} & \textbf{Overall F1} & \textbf{Politics} & \textbf{Pornography} & \textbf{Violence} & \textbf{Gambling} & \textbf{Bias} & \textbf{Harmless} \\
\midrule
w/o k-NN Retrieval              & 86.9 (-2.3) & 85.4 (-3.9) & 69.0 (-2.5) & 96.6 (-1.2) & 78.5 (-3.5) & 93.6 (-2.5) & 98.2 (-0.4) \\
w/o Second-stage SFT       & 88.0 (-1.2) & 86.1 (-3.2) & 69.3 (-2.2) & 97.2 (-0.6) & 81.2 (-0.8) & 95.8 (-0.3) & 98.4 (-0.2) \\
\textbf{\chairo (Ours)}    & \textbf{89.2} & \textbf{89.3} & \textbf{71.5} & \textbf{97.8} & \textbf{82.0} & \textbf{96.1} & \textbf{98.6} \\
\bottomrule
\end{tabular}
\caption{Ablation Study on Model Components with Performance Change}
\label{tab:ablation_comparison_updated}
\end{table*}

\begin{table}[htbp]
\centering
\caption{F1-Score Comparison on Aegis Dataset}
\label{tab:common_categories_f1}
\footnotesize
\setlength{\tabcolsep}{2pt}   % 列间距再压缩
\renewcommand{\arraystretch}{1.05} % 行高微调
\begin{tabular}{@{}lcccc@{}} % @{} 去掉两侧空隙
\toprule
\textbf{Categories} & \textbf{Qwen3-8B} & \textbf{RAG} & \textbf{SFT} & \textbf{\chairo(Ours)} \\
\midrule
Hate          & 67.2 & 62.3 & 74.7 & \textbf{81.5} \\
Sexual        & 65.2 & 72.3 & 71.4 & \textbf{81.6} \\
Confessions   & 66.4 & 68.3 & 80.0 & \textbf{82.5} \\
Harassment    & 25.8 & 27.6 & 26.7 & \textbf{40.7} \\
Profanity     & 40.0 & 42.9 & \textbf{55.0} & 53.9 \\
\midrule
Average F1-Score & 52.9 & 54.7 & 61.6 & \textbf{68.0} \\
\bottomrule
\end{tabular}
\vspace{1em}
\end{table}

% \begin{table}[htbp]
% \centering
% \scriptsize
% \begin{tabular}{lccccccc}
% \toprule
% \textbf{Model} & \textbf{Average} & \textbf{Politics} & \textbf{Pornography} & \textbf{Violence} & \textbf{Gambling} & \textbf{Bias} & \textbf{Harmless} \\
% \midrule
% Qwen2.5-32B-Instruct & 74.3 & 59.1 & 91.1 & 84.4 & 95.4 & 67.9 & 54.2 \\
% Simple Rule & 75.1 & 70.5 & 91.1 & 80.7 & 93.5 & 59.7 & 55.0 \\
% RQ4 & 88.7 & 88.7 & 95.7 & 97.4 & 97.8 & 82.2 & 70.6 \\
% \bottomrule
% \end{tabular}
% \end{table}
\begin{table}[t]
\centering
\begin{tabular}{lcc}
\toprule
\textbf{Model}                  & \textbf{F1} & \textbf{Human (\%)} \\
\midrule
Qwen2.5-32B-Instruct            & 74.3        & -                   \\
Simple Rule                     & 75.1        & 15                  \\
RQ4                             & 88.7        & 85                  \\
\bottomrule
\end{tabular}
\caption{Rule Quality Evaluation Results. The F1-score reflects the generalization ability of rules when applied to an external moderation model. The "Human (\%)" denotes the preference rate derived from a double-blind comparison of 100 test cases by three annotators with content moderation experience, where rules are assessed for contextual relevance, completeness, and alignment with human judgment criteria.}
\label{tab:rule_quality}
\end{table}

\subsection{Discussion}
The experimental results presented in the preceding section demonstrate that our proposed \chairo framework, by integrating analogical reasoning, explicit rule induction, and hierarchical reasoning chain optimization, significantly enhances the performance of LLM-based content moderation systems across various content moderation tasks. In this section, we delve deeper into the contributions of individual components, provide case studies for qualitative insights, and present human evaluations of the generated rule quality to further elucidate the framework's effectiveness and practical implications.

\subsubsection{Ablation Study}

To isolate the contributions of key components in our \chairo framework, we conduct ablation experiments by systematically removing core modules and measuring the resulting performance degradation, as reported in Table \ref{tab:ablation_comparison_updated}.

\textbf{Role of the Retrieval Module.} In the ``w/o k-NN Retrieval'' setting, we replace the k-NN-based analogical retrieval in Stage~1 with random sample selection. This leads to a 2.3\% drop in the overall F1 score, with particularly notable declines in politically sensitive content (-3.9\%) and gambling-related content (-3.5\%). The result underscores the critical role of retrieving relevant analogical cases in capturing subtle moderation criteria, especially in domains where rules are implicit or context-dependent. Without the most relevant analogical examples, the model struggles to generalize beyond superficial patterns, often leading to oversimplified judgments.

\textbf{Importance of the Second-Stage Fine-Tuning.} In the ``w/o Second-stage SFT'' setting, we skip the rule injection and second-round fine-tuning in Stage~3, so the model relies solely on Stage~1 analogical reasoning. This leads to a 1.2\% decline in overall F1 score, with the most significant drops observed in political content (-3.2\%) and pornography (-2.2\%). The result highlights the value of combining explicit rules with analogical reasoning. The second-stage fine-tuning ensures that rules are dynamically adapted to specific content contexts, bridging the gap between general guidelines and case-specific nuances.

These results confirm that the retrieval and rule injection components complement each other: retrieval supplies relevant examples, and rule induction turns them into generalizable moderation standards.

\subsubsection{Results on Other Dataset}
We further evaluate model performance on additional datasets to assess the cross-dataset generalization and harmful content recognition capabilities of each method. As summarized in Table \ref{tab:common_categories_f1}, there are clear differences in F1 scores across categories and models. Our proposed method, \chairo  achieves the highest overall average F1 score of 68.0, outperforming SFT (61.6), Qwen3-8B (52.9), and RAG (54.7).

A closer look at category-level results reveals that \chairo consistently achieves the best or second-best F1 scores in key harmful content categories, including "Hate" (81.5), "Sexual" (81.6), "Confessions" (82.5), "Harassment" (40.7), and "Profanity" (53.9). Notably, the improvements in the "Hate" and "Sexual" categories are particularly pronounced compared to other methods.

Overall, these results demonstrate that \chairo exhibits stronger and more comprehensive performance in most harmful content recognition categories, validating its effectiveness and generalizability in cross-dataset scenarios. This further highlights the practical value of our approach for robust and reliable harmful content detection in diverse real-world settings.
\subsubsection{Rule Quality Evaluation}
  How effective are the moderation rules generated by our proposed method in practical moderation scenarios?

 To address Research Question 4, we evaluate the quality of the moderation rules themselves, beyond end-to-end moderation accuracy. We compare rules generated by our analogical approach against those produced by single-instance LLM prompting, using two complementary evaluations summarized in Table~\ref{tab:rule_quality}.

\textbf{Human Evaluation.} We invited three annotators with practical experience in content moderation to independently rate the quality, clarity, and usefulness of the moderation rules produced by our method and the baseline. Each annotator independently reviewed 100 randomly ordered pairs of rules from different methods in a double-blind setting, and selected the one they judged more contextually relevant, complete, and reliable. The annotators consistently preferred rules generated by our analogical approach, highlighting their clarity and practical utility. Across the 100 test cases, the three annotators preferred our rules in 85\% of cases, compared to only 15\% for the simple rule-based baseline. The Qwen2.5-32B-Instruct baseline was not included in the human preference comparison. These results indicate that analogically generated rules align more closely with human moderation judgment.

\textbf{Generalization Assessment with External Models.} We further evaluated the generalizability of the generated rules by injecting them into an external model distinct from our base model Qwen3-8B. Rules generated by our method achieve an F1-score of 88.7 when applied to the external model, outperforming both the Qwen2.5-32B-Instruct baseline (F1 = 74.3) and the simple rule-based approach (F1 = 75.1). This represents an absolute improvement of 14.4 and 13.6 percentage points respectively, confirming that the generated rules transfer well beyond the original modeling context.

\subsubsection{Adaptability to Evolving Standards}
\label{sec:adaptability}

Moderation standards shift over time and differ across cultures and platforms. The out-of-distribution results in Table~\ref{tab:common_categories_f1} show that \chairo retains strong performance under distributional shift, suggesting that the combination of analogical reasoning and explicit rules provides robustness beyond the training distribution. Moreover, the modular design of our framework supports lightweight fine-tuning on recent data when the shift is moderate, without requiring a full retraining of the pipeline.

\subsubsection{On the Choice of Supervised Fine-Tuning}
\label{sec:rl_discussion}

We explored reinforcement learning as an alternative to SFT in preliminary experiments. However, binary reward signals based on label correctness proved too coarse for this task, and the model quickly learned superficial shortcuts instead of performing genuine analogical reasoning and rule application. We attribute this to the multi-step nature of our reasoning chain, where intermediate quality matters but is not captured by a single binary reward. Designing reward functions that also account for reasoning chain quality is a promising direction for future work.

\section{Related Work}
\subsection{LLMs for Content Moderation}
In recent years, LLMs have garnered significant attention in content moderation research due to their strong natural language understanding capabilities. Existing LLM-based moderation approaches can be broadly categorized into two paradigms:

\textbf{Prompt Engineering-driven Direct Moderation}: Models such as GPT-series can directly generate moderation decisions from carefully designed prompts~\citep{radford2019language}. This approach does not require large-scale annotated data and can inherently provide detailed reasoning processes, enhancing interpretability~\citep{zhan2024slm, Li2025demod}. However, when moderation criteria are inherently ambiguous, model predictions become sensitive to subtle variations in prompt wording, leading to inconsistent moderation outcomes~\citep{rottger2021two, kristina2024counter}.

\textbf{Fine-tuning-based Domain Adaptation}: Post-training large pretrained models on annotated moderation datasets has been explored extensively. For example, contrastive fine-tuning approaches leverage labeled datasets containing both compliant and violating content to enhance models' sensitivity to specific moderation rules~\citep{devlin2019bert}. While fine-tuning clearly improves performance on explicitly annotated rules, it suffers from data-dependency issues: high-quality annotated moderation datasets are costly to acquire, and fine-tuned models often fail to generalize to previously unseen moderation rules or subtle semantic nuances~\citep{jha2024memeguard, hakan2024llamaguard}.

\subsection{Generation and Optimization of Moderation Rules}
The explicitness and clarity of moderation rules are critical for ensuring reliable moderation decisions. However, existing methods for constructing moderation rules have notable shortcomings:

\textbf{Manually-defined High-level Rules}: Current moderation systems commonly utilize abstract categories like hate speech and adult content as moderation guidelines. However, such broad categories often fail to adequately cover fine-grained real-world scenarios. For instance, subtle forms of harassment or discrimination expressed through metaphor, euphemism, or implicit linguistic cues cannot be effectively captured by simple keyword-based rules~\citep{mei2024hiddenguard, Wang2024moderator, palla2025policy}. Furthermore, exhaustively enumerating all possible instances or patterns of problematic content through manual effort alone is practically infeasible.

\textbf{LLM-driven Adaptive Rule Generation}: Recent studies have explored prompting-based approaches that use LLMs to automatically generate moderation rules by summarizing characteristics of violating content~\citep{franco2023analyzing}. However, this strategy relies heavily on the generic world knowledge encoded in large models and often neglects domain-specific expert knowledge such as platform-specific community guidelines. Additionally, current LLM-driven rule generation processes are typically independent from the moderation decision-making stage. As a consequence, generated rules lack the flexibility and context-awareness required to dynamically adapt to specific content instances, limiting the practical effectiveness of such approaches in realistic moderation scenarios~\citep{masud2024hate}.

\section{Conclusion}

We presented \chairo, a framework that jointly optimizes analogical retrieval, rule induction, and hierarchical reasoning for LLM-based content moderation. On both in-distribution and out-of-distribution benchmarks, \chairo outperforms SFT and RAG baselines by a clear margin. Human annotators also preferred the rules induced by our method over single-instance alternatives in 85\% of cases. The modular design also makes each moderation decision traceable to explicit rules and analogical evidence, improving interpretability for practical deployment.

\section*{Limitations}
Although this study aims to provide a reliable and robust moderation approach for harmful content on real-world online platforms, several important limitations remain.

First, this study primarily focuses on the textual modality and has not yet extended the proposed reasoning paradigm to multimodal large language models. In the contemporary context where audiovisual content is increasingly prevalent, moderation of multimodal content, including short videos and live streaming, is an equally pressing challenge. A key direction for future work is to extend the proposed paradigm to content moderation tasks involving multimodal data.

In addition, although this study conducts systematic experimental analyses across several real-world datasets, the effectiveness and robustness of the model in actual platforms and complex application scenarios remain to be further validated, particularly in contexts involving interactive dialogues, contextual dependencies, and user diversity.

Additionally, while our framework shows robustness to moderate distributional shifts, drastic changes in moderation standards across time or culture may still require retraining or rule reconstruction (see Section~\ref{sec:adaptability} for further discussion). Similarly, extending beyond supervised fine-tuning to reinforcement learning remains an open challenge, as discussed in Section~\ref{sec:rl_discussion}.

\section*{Acknowledgments}
This work was supported by the National Natural Science Funds for Young Scholar under Grant 62503336.

\section*{Ethical considerations}
\textbf{Exposure to Offensive Content:}
During this study, we encountered and curated a substantial amount of offensive content for the purpose of constructing the research dataset. All authors were fully aware of the nature of the study and consented to review such materials. It is important to note that, owing to the use of an isolated experimental protocol, no individuals other than the authors were exposed to these materials.

\bibliographystyle{unsrtnat}
\bibliography{custom}

@inproceedings{yuan2024rigorllm,
  title={RigorLLM: resilient guardrails for large language models against undesired content},
  author={Yuan, Zhuowen and Xiong, Zidi and Zeng, Yi and Yu, Ning and Jia, Ruoxi and Song, Dawn and Li, Bo},
  booktitle={Proceedings of the 41st International Conference on Machine Learning},
  pages={57953--57965},
  year={2024}
}

@misc{zeng2024shieldgemma,
      title={ShieldGemma: Generative AI Content Moderation Based on Gemma}, 
      author={Wenjun Zeng and Yuchi Liu and Ryan Mullins and Ludovic Peran and Joe Fernandez and Hamza Harkous and Karthik Narasimhan and Drew Proud and Piyush Kumar and Bhaktipriya Radharapu and Olivia Sturman and Oscar Wahltinez},
      year={2024},
      eprint={2407.21772},
      archivePrefix={arXiv},
      primaryClass={cs.CL},
      url={https://arxiv.org/abs/2407.21772}, 
}

@inproceedings{kolla2024llm,
  title={Llm-mod: Can large language models assist content moderation?},
  author={Kolla, Mahi and Salunkhe, Siddharth and Chandrasekharan, Eshwar and Saha, Koustuv},
  booktitle={Extended Abstracts of the CHI Conference on Human Factors in Computing Systems},
  pages={1--8},
  year={2024}
}

@article{ouyang2022training,
  title={Training language models to follow instructions with human feedback},
  author={Ouyang, Long and Wu, Jeffrey and Jiang, Xu and Almeida, Diogo and Wainwright, Carroll and Mishkin, Pamela and Zhang, Chong and Agarwal, Sandhini and Slama, Katarina and Ray, Alex and others},
  journal={Advances in neural information processing systems},
  volume={35},
  pages={27730--27744},
  year={2022}
}

@article{rafailov2023direct,
  title={Direct preference optimization: Your language model is secretly a reward model},
  author={Rafailov, Rafael and Sharma, Archit and Mitchell, Eric and Manning, Christopher D and Ermon, Stefano and Finn, Chelsea},
  journal={Advances in Neural Information Processing Systems},
  volume={36},
  pages={53728--53741},
  year={2023}
}

@inproceedings{khaliq2024ragar,
    title = "{RAGAR}, Your Falsehood Radar: {RAG}-Augmented Reasoning for Political Fact-Checking using Multimodal Large Language Models",
    author = "Khaliq, Mohammed Abdul  and
      Chang, Paul Yu-Chun  and
      Ma, Mingyang  and
      Pflugfelder, Bernhard  and
      Mileti{\'c}, Filip",
    booktitle = "Proceedings of the Seventh Fact Extraction and VERification Workshop (FEVER)",
    month = nov,
    year = "2024",
    address = "Miami, Florida, USA",
    publisher = "Association for Computational Linguistics",
    url = "https://aclanthology.org/2024.fever-1.29/",
    doi = "10.18653/v1/2024.fever-1.29",
    pages = "280--296",
}

@misc{liu2025guardreasoner,
      title={GuardReasoner: Towards Reasoning-based LLM Safeguards}, 
      author={Yue Liu and Hongcheng Gao and Shengfang Zhai and Jun Xia and Tianyi Wu and Zhiwei Xue and Yulin Chen and Kenji Kawaguchi and Jiaheng Zhang and Bryan Hooi},
      year={2025},
      eprint={2501.18492},
      archivePrefix={arXiv},
      primaryClass={cs.CR},
      url={https://arxiv.org/abs/2501.18492}, 
}

@misc{ma2023adapting,
      title={Adapting Large Language Models for Content Moderation: Pitfalls in Data Engineering and Supervised Fine-tuning}, 
      author={Huan Ma and Changqing Zhang and Huazhu Fu and Peilin Zhao and Bingzhe Wu},
      year={2024},
      eprint={2310.03400},
      archivePrefix={arXiv},
      primaryClass={cs.LG},
      url={https://arxiv.org/abs/2310.03400}, 
}

@article{radford2019language,
  title={Language models are unsupervised multitask learners},
  author={Radford, Alec and Wu, Jeffrey and Child, Rewon and Luan, David and Amodei, Dario and Sutskever, Ilya and others},
  journal={OpenAI blog},
  volume={1},
  number={8},
  pages={9},
  year={2019}
}

@article{brown2020language,
  title={Language models are few-shot learners},
  author={Brown, Tom and Mann, Benjamin and Ryder, Nick and Subbiah, Melanie and Kaplan, Jared D and Dhariwal, Prafulla and Neelakantan, Arvind and Shyam, Pranav and Sastry, Girish and Askell, Amanda and others},
  journal={Advances in neural information processing systems},
  volume={33},
  pages={1877--1901},
  year={2020}
}

@inproceedings{palla2025policy,
  author       = {Konstantina Palla and
                  Jos{\'{e}} Luis Redondo Garc{\'{\i}}a and
                  Claudia Hauff and
                  Francesco Fabbri and
                  Andreas Damianou and
                  Henrik Lindstr{\"{o}}m and
                  Daniel R. Taber and
                  Mounia Lalmas},
  title        = {Policy-as-Prompt: Rethinking Content Moderation in the Age of Large
                  Language Models},
  booktitle    = {Proceedings of the 2025 {ACM} Conference on Fairness, Accountability,
                  and Transparency, FAccT 2025, Athens, Greece, June 23-26, 2025},
  pages        = {840--854},
  publisher    = {{ACM}},
  year         = {2025},
  url          = {https://doi.org/10.1145/3715275.3732054},
  doi          = {10.1145/3715275.3732054},
  bibsource    = {dblp computer science bibliography, https://dblp.org}
}

@inproceedings{kumar2024watch,
  title={Watch your language: Investigating content moderation with large language models},
  author={Kumar, Deepak and AbuHashem, Yousef Anees and Durumeric, Zakir},
  booktitle={Proceedings of the International AAAI Conference on Web and Social Media},
  volume={18},
  pages={865--878},
  year={2024}
}

@inproceedings{masud2024hate,
  author       = {Sarah Masud and
                  Sahajpreet Singh and
                  Viktor Hangya and
                  Alexander Fraser and
                  Tanmoy Chakraborty},
  editor       = {Yaser Al{-}Onaizan and
                  Mohit Bansal and
                  Yun{-}Nung Chen},
  title        = {Hate Personified: Investigating the role of LLMs in content moderation},
  booktitle    = {Proceedings of the 2024 Conference on Empirical Methods in Natural
                  Language Processing, {EMNLP} 2024, Miami, FL, USA, November 12-16,
                  2024},
  pages        = {15847--15863},
  publisher    = {Association for Computational Linguistics},
  year         = {2024},
  url          = {https://doi.org/10.18653/v1/2024.emnlp-main.886},
  doi          = {10.18653/V1/2024.EMNLP-MAIN.886},
  bibsource    = {dblp computer science bibliography, https://dblp.org}
}

@article{chandrasekharan2019crossmod,
  title={Crossmod: A cross-community learning-based system to assist reddit moderators},
  author={Chandrasekharan, Eshwar and Gandhi, Chaitrali and Mustelier, Matthew Wortley and Gilbert, Eric},
  journal={Proceedings of the ACM on human-computer interaction},
  volume={3},
  number={CSCW},
  pages={1--30},
  year={2019},
  publisher={ACM New York, NY, USA}
}

@inproceedings{zhan2024slm,
  author       = {Xianyang Zhan and
                  Agam Goyal and
                  Yilun Chen and
                  Eshwar Chandrasekharan and
                  Koustuv Saha},
  editor       = {Luis Chiruzzo and
                  Alan Ritter and
                  Lu Wang},
  title        = {SLM-Mod: Small Language Models Surpass LLMs at Content Moderation},
  booktitle    = {Proceedings of the 2025 Conference of the Nations of the Americas
                  Chapter of the Association for Computational Linguistics: Human Language
                  Technologies, {NAACL} 2025 - Volume 1: Long Papers, Albuquerque, New
                  Mexico, USA, April 29 - May 4, 2025},
  pages        = {8774--8790},
  publisher    = {Association for Computational Linguistics},
  year         = {2025},
  url          = {https://doi.org/10.18653/v1/2025.naacl-long.441},
  doi          = {10.18653/V1/2025.NAACL-LONG.441},
  bibsource    = {dblp computer science bibliography, https://dblp.org}
}

@inproceedings{rottger2021two,
    title = "Two Contrasting Data Annotation Paradigms for Subjective {NLP} Tasks",
    author = {R{\"o}ttger, Paul  and
      Vidgen, Bertie  and
      Hovy, Dirk  and
      Pierrehumbert, Janet},
    booktitle = "Proceedings of the 2022 Conference of the North American Chapter of the Association for Computational Linguistics: Human Language Technologies",
    month = jul,
    year = "2022",
    address = "Seattle, United States",
    publisher = "Association for Computational Linguistics",
    url = "https://aclanthology.org/2022.naacl-main.13/",
    doi = "10.18653/v1/2022.naacl-main.13",
    pages = "175--190",
}

@inproceedings{devlin2019bert,
  title={Bert: Pre-training of deep bidirectional transformers for language understanding},
  author={Devlin, Jacob and Chang, Ming-Wei and Lee, Kenton and Toutanova, Kristina},
  booktitle={Proceedings of the 2019 conference of the North American chapter of the association for computational linguistics: human language technologies, volume 1 (long and short papers)},
  pages={4171--4186},
  year={2019}
}

@inproceedings{jha2024memeguard,
    title = "{M}eme{G}uard: An {LLM} and {VLM}-based Framework for Advancing Content Moderation via Meme Intervention",
    author = "Jha, Prince  and
      Jain, Raghav  and
      Mandal, Konika  and
      Chadha, Aman  and
      Saha, Sriparna  and
      Bhattacharyya, Pushpak",
    booktitle = "Proceedings of the 62nd Annual Meeting of the Association for Computational Linguistics (Volume 1: Long Papers)",
    month = aug,
    year = "2024",
    address = "Bangkok, Thailand",
    publisher = "Association for Computational Linguistics",
    url = "https://aclanthology.org/2024.acl-long.439/",
    doi = "10.18653/v1/2024.acl-long.439",
    pages = "8084--8104",
}

@inproceedings{franco2023analyzing,
  title={Analyzing the use of large language models for content moderation with chatgpt examples},
  author={Franco, Mirko and Gaggi, Ombretta and Palazzi, Claudio E},
  booktitle={Proceedings of the 3rd International Workshop on Open Challenges in Online Social Networks},
  pages={1--8},
  year={2023}
}

@inproceedings{zheng2024llamafactory,
    title = "{L}lama{F}actory: Unified Efficient Fine-Tuning of 100+ Language Models",
    author = "Zheng, Yaowei  and
      Zhang, Richong  and
      Zhang, Junhao  and
      Ye, Yanhan  and
      Luo, Zheyan",
    booktitle = "Proceedings of the 62nd Annual Meeting of the Association for Computational Linguistics (Volume 3: System Demonstrations)",
    month = aug,
    year = "2024",
    address = "Bangkok, Thailand",
    publisher = "Association for Computational Linguistics",
    url = "https://aclanthology.org/2024.acl-demos.38/",
    doi = "10.18653/v1/2024.acl-demos.38",
    pages = "400--410",
}

@inproceedings{rajbhandari2020zero,
  title={Zero: Memory optimizations toward training trillion parameter models},
  author={Rajbhandari, Samyam and Rasley, Jeff and Ruwase, Olatunji and He, Yuxiong},
  booktitle={SC20: International Conference for High Performance Computing, Networking, Storage and Analysis},
  pages={1--16},
  year={2020},
  organization={IEEE}
}

@misc{micikevicius2017mixed,
      title={Mixed Precision Training}, 
      author={Paulius Micikevicius and Sharan Narang and Jonah Alben and Gregory Diamos and Erich Elsen and David Garcia and Boris Ginsburg and Michael Houston and Oleksii Kuchaiev and Ganesh Venkatesh and Hao Wu},
      year={2018},
      eprint={1710.03740},
      archivePrefix={arXiv},
      primaryClass={cs.AI},
      url={https://arxiv.org/abs/1710.03740}, 
}

@inproceedings{fan2018hierarchical,
    title = "Hierarchical Neural Story Generation",
    author = "Fan, Angela  and
      Lewis, Mike  and
      Dauphin, Yann",
    booktitle = "Proceedings of the 56th Annual Meeting of the Association for Computational Linguistics (Volume 1: Long Papers)",
    month = jul,
    year = "2018",
    address = "Melbourne, Australia",
    publisher = "Association for Computational Linguistics",
    url = "https://aclanthology.org/P18-1082/",
    doi = "10.18653/v1/P18-1082",
    pages = "889--898",
}

@misc{holtzman2019curious,
      title={The Curious Case of Neural Text Degeneration}, 
      author={Ari Holtzman and Jan Buys and Li Du and Maxwell Forbes and Yejin Choi},
      year={2020},
      eprint={1904.09751},
      archivePrefix={arXiv},
      primaryClass={cs.CL},
      url={https://arxiv.org/abs/1904.09751}, 
}

@article{huang2025content,
  title={Content moderation by llm: From accuracy to legitimacy},
  author={Huang, Tao},
  journal={Artificial Intelligence Review},
  volume={58},
  number={10},
  pages={1--32},
  year={2025},
  publisher={Springer}
}

@inproceedings{keluskar2024llms,
  title={Do llms understand ambiguity in text? a case study in open-world question answering},
  author={Keluskar, Aryan and Bhattacharjee, Amrita and Liu, Huan},
  booktitle={2024 IEEE International Conference on Big Data (BigData)},
  pages={7485--7490},
  year={2024},
  organization={IEEE}
}

@inproceedings{he2024cpl,
  title={Cpl-novid: Context-aware prompt-based learning for norm violation detection in online communities},
  author={He, Zihao and May, Jonathan and Lerman, Kristina},
  booktitle={Proceedings of the International AAAI Conference on Web and Social Media},
  volume={18},
  pages={569--582},
  year={2024}
}

@misc{chen2024class,
      title={Class-RAG: Real-Time Content Moderation with Retrieval Augmented Generation}, 
      author={Jianfa Chen and Emily Shen and Trupti Bavalatti and Xiaowen Lin and Yongkai Wang and Shuming Hu and Harihar Subramanyam and Ksheeraj Sai Vepuri and Ming Jiang and Ji Qi and Li Chen and Nan Jiang and Ankit Jain},
      year={2024},
      eprint={2410.14881},
      archivePrefix={arXiv},
      primaryClass={cs.AI},
      url={https://arxiv.org/abs/2410.14881}, 
}

@inproceedings{kristina2024counter,
  author       = {Kristina Gligoric and
                  Myra Cheng and
                  Lucia Zheng and
                  Esin Durmus and
                  Dan Jurafsky},
  title        = {{NLP} Systems That Can't Tell Use from Mention Censor Counterspeech,
                  but Teaching the Distinction Helps},
  booktitle    = {Proceedings of the 2024 Conference of the North American Chapter of
                  the Association for Computational Linguistics: Human Language Technologies
                  (Volume 1: Long Papers), {NAACL} 2024, Mexico City, Mexico, June 16-21,
                  2024},
  pages        = {5942--5959},
  year         = {2024},
}

@article{Yang2025qwen3,
  author       = {An Yang and
                  Anfeng Li and
                  Baosong Yang and
                  Beichen Zhang and
                  Binyuan Hui and
                  Bo Zheng and
                  Bowen Yu and
                  Chang Gao and
                  Chengen Huang and
                  Chenxu Lv and
                  Chujie Zheng and
                  Dayiheng Liu and
                  Fan Zhou and
                  Fei Huang and
                  Feng Hu and
                  Hao Ge and
                  Haoran Wei and
                  Huan Lin and
                  Jialong Tang and
                  Jian Yang and
                  Jianhong Tu and
                  Jianwei Zhang and
                  Jian Yang and
                  Jiaxi Yang and
                  Jingren Zhou and
                  Jingren Zhou and
                  Junyang Lin and
                  Kai Dang and
                  Keqin Bao and
                  Kexin Yang and
                  Le Yu and
                  Lianghao Deng and
                  Mei Li and
                  Mingfeng Xue and
                  Mingze Li and
                  Pei Zhang and
                  Peng Wang and
                  Qin Zhu and
                  Rui Men and
                  Ruize Gao and
                  Shixuan Liu and
                  Shuang Luo and
                  Tianhao Li and
                  Tianyi Tang and
                  Wenbiao Yin and
                  Xingzhang Ren and
                  Xinyu Wang and
                  Xinyu Zhang and
                  Xuancheng Ren and
                  Yang Fan and
                  Yang Su and
                  Yichang Zhang and
                  Yinger Zhang and
                  Yu Wan and
                  Yuqiong Liu and
                  Zekun Wang and
                  Zeyu Cui and
                  Zhenru Zhang and
                  Zhipeng Zhou and
                  Zihan Qiu},
  title        = {Qwen3 Technical Report},
  journal      = {CoRR},
  volume       = {abs/2505.09388},
  year         = {2025},
}

@article{hakan2024llamaguard,
  author       = {Hakan Inan and
                  Kartikeya Upasani and
                  Jianfeng Chi and
                  Rashi Rungta and
                  Krithika Iyer and
                  Yuning Mao and
                  Michael Tontchev and
                  Qing Hu and
                  Brian Fuller and
                  Davide Testuggine and
                  Madian Khabsa},
  title        = {Llama Guard: LLM-based Input-Output Safeguard for Human-AI Conversations},
  journal      = {CoRR},
  volume       = {abs/2312.06674},
  year         = {2023},
}

@article{mei2024hiddenguard,
  author       = {Lingrui Mei and
                  Shenghua Liu and
                  Yiwei Wang and
                  Baolong Bi and
                  Ruibin Yuan and
                  Xueqi Cheng},
  title        = {HiddenGuard: Fine-Grained Safe Generation with Specialized Representation
                  Router},
  journal      = {CoRR},
  volume       = {abs/2410.02684},
  year         = {2024},
}

@article{Li2025demod,
  author       = {Yaqiong Li and
                  Peng Zhang and
                  Hansu Gu and
                  Tun Lu and
                  Siyuan Qiao and
                  Yubo Shu and
                  Yiyang Shao and
                  Ning Gu},
  title        = {DeMod: {A} Holistic Tool with Explainable Detection and Personalized
                  Modification for Toxicity Censorship},
  journal      = {Proc. {ACM} Hum. Comput. Interact.},
  volume       = {9},
  number       = {2},
  pages        = {1--24},
  year         = {2025},
}

@inproceedings{RebedeaDSPC23NeMo,
  author       = {Traian Rebedea and
                  Razvan Dinu and
                  Makesh Narsimhan Sreedhar and
                  Christopher Parisien and
                  Jonathan Cohen},
  editor       = {Yansong Feng and
                  Els Lefever},
  title        = {NeMo Guardrails: {A} Toolkit for Controllable and Safe {LLM} Applications
                  with Programmable Rails},
  booktitle    = {Proceedings of the 2023 Conference on Empirical Methods in Natural
                  Language Processing, {EMNLP} 2023 - System Demonstrations, Singapore,
                  December 6-10, 2023},
  pages        = {431--445},
  year         = {2023},
}

@inproceedings{Wu25ICM,
  author       = {Mengyang Wu and
                  Yuzhi Zhao and
                  Jialun Cao and
                  Mingjie Xu and
                  Zhongming Jiang and
                  Xuehui Wang and
                  Qinbin Li and
                  Guangneng Hu and
                  Shengchao Qin and
                  Chi{-}Wing Fu},
  title        = {ICM-Assistant: Instruction-tuning Multimodal Large Language Models
                  for Rule-based Explainable Image Content Moderation},
  booktitle    = {AAAI-25, Sponsored by the Association for the Advancement of Artificial
                  Intelligence, February 25 - March 4, 2025, Philadelphia, PA, {USA}},
  pages        = {8413--8422},
  year         = {2025},
}

@inproceedings{Nishant24CoT,
  author       = {Nishant Vishwamitra and
                  Keyan Guo and
                  Farhan Tajwar Romit and
                  Isabelle Ondracek and
                  Long Cheng and
                  Ziming Zhao and
                  Hongxin Hu},
  title        = {Moderating New Waves of Online Hate with Chain-of-Thought Reasoning
                  in Large Language Models},
  booktitle    = {{IEEE} Symposium on Security and Privacy, {SP} 2024, San Francisco,
                  CA, USA, May 19-23, 2024},
  pages        = {788--806},
  publisher    = {{IEEE}},
  year         = {2024},
  url          = {https://doi.org/10.1109/SP54263.2024.00181},
  doi          = {10.1109/SP54263.2024.00181},
  bibsource    = {dblp computer science bibliography, https://dblp.org}
}

@inproceedings{Wang2024moderator,
  author       = {Peiran Wang and
                  Qiyu Li and
                  Longxuan Yu and
                  Ziyao Wang and
                  Ang Li and
                  Haojian Jin},
  title        = {Moderator: Moderating Text-to-Image Diffusion Models through Fine-grained
                  Context-based Policies},
  booktitle    = {Proceedings of the 2024 on {ACM} {SIGSAC} Conference on Computer and
                  Communications Security, {CCS} 2024, Salt Lake City, UT, USA, October
                  14-18, 2024},
  pages        = {1181--1195},
  year         = {2024},
}

@inproceedings{NghiemD24HateCOT,
  author       = {Huy Nghiem and
                  Hal Daum{\'{e}} III},
  title        = {HateCOT: An Explanation-Enhanced Dataset for Generalizable Offensive
                  Speech Detection via Large Language Models},
  booktitle    = {Findings of the Association for Computational Linguistics: {EMNLP}
                  2024, Miami, Florida, USA, November 12-16, 2024},
  pages        = {5938--5956},
  year         = {2024},
}

@inproceedings{Wu24Legilimens,
  author       = {Jialin Wu and
                  Jiangyi Deng and
                  Shengyuan Pang and
                  Yanjiao Chen and
                  Jiayang Xu and
                  Xinfeng Li and
                  Wenyuan Xu},
  title        = {Legilimens: Practical and Unified Content Moderation for Large Language
                  Model Services},
  booktitle    = {Proceedings of the 2024 on {ACM} {SIGSAC} Conference on Computer and
                  Communications Security, {CCS} 2024, Salt Lake City, UT, USA, October
                  14-18, 2024},
  pages        = {1151--1165},
  year         = {2024},
}

@misc{bge-m3,
      title={BGE M3-Embedding: Multi-Lingual, Multi-Functionality, Multi-Granularity Text Embeddings Through Self-Knowledge Distillation}, 
      author={Jianlv Chen and Shitao Xiao and Peitian Zhang and Kun Luo and Defu Lian and Zheng Liu},
      year={2024},
      eprint={2402.03216},
      archivePrefix={arXiv},
      primaryClass={cs.CL}
}

\appendix

\section{Prompts}
\begin{figure*}[h]
\centering
\includegraphics[width=1.0\textwidth]{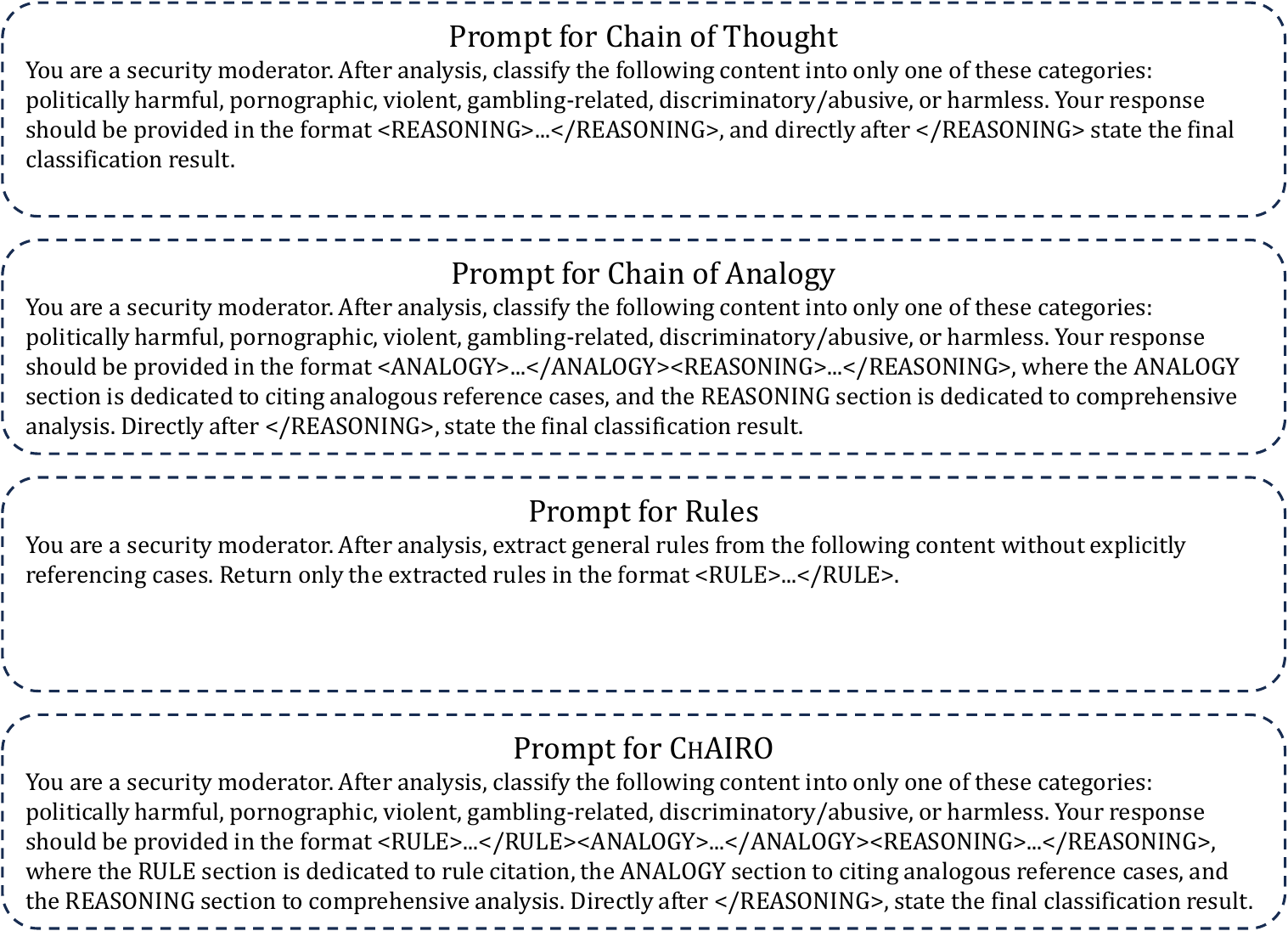}
\caption{Prompt for synthesizing the chain of analogical inductive reasoning.}
\label{fig:prompt_chains}
\end{figure*}
We modularly synthesize each constituent of the proposed specialized chain-of-thought, thereby both transparently elucidating the contribution of every component to the overall reasoning trajectory and progressively enhancing the chain's logical coherence and robustness through iterative leverage of a more capable large language model, as shown in Figure \ref{fig:prompt_chains}.
\begin{itemize}
    \item Chain of Thought:
    In this stage, we generate a chain-of-thought-style reasoning trajectory for each input instance, thereby furnishing the foundational inferential logic that underpins both the focal sample and its reference exemplars.
    \item Chain of Analogy:
    In this stage, we augment existing labeled moderation data through a bootstrapping retrieval-enhancement procedure,  thereby endowing the model with an analogical-reasoning epistemic modality.
    \item Induction Rules:
    In this stage, we perform explicit rule induction to extract moderation rules from analogous examples generated by the previously trained analogical reasoning mode, thereby furnishing a principled data substrate for subsequent inductive-capability infusion.
    \item \chairo:
    In this stage, we inject the moderation rules derived
from the previous rule induction step back into the reasoning process to further enhance the moderation capabilities of the model.
\end{itemize}

\end{document}